\documentclass{article}
\usepackage{arxiv}
\usepackage{enumitem}

\usepackage[T1]{fontenc}
\usepackage[utf8]{inputenc}

\usepackage[table]{xcolor}
\usepackage{graphicx}
\graphicspath{{./}{./images/}}   % 根目录与 images/ 都可搜到

\usepackage{booktabs}
\usepackage{array}
\usepackage{siunitx}
\usepackage{float}
\usepackage{microtype}
\usepackage{nicefrac}

\usepackage{amsfonts}
\usepackage{amsmath}
\usepackage{caption}
\usepackage{algorithm}
\usepackage{algpseudocode} % from algorithmicx
\algrenewcommand\algorithmicrequire{\textbf{Inputs:}}
\algrenewcommand\algorithmicensure{\textbf{Outputs:}}
\algrenewcommand\algorithmiccomment[1]{\hfill$\triangleright$~#1}
\DeclareMathOperator*{\argmin}{arg\,min}

\usepackage{url}
\usepackage{hyperref}

\definecolor{headgreen}{RGB}{232,245,233}
\definecolor{extrow}{gray}{0.96}
\usepackage{tabularx}   % 自适应表宽
\usepackage{xurl}       % \url{} 可在更多位置断行（含下划线）
\usepackage{seqsplit}   % 超长连续字符串按字符断行
\usepackage{placeins} % 控制浮动体不越节
\usepackage{url}
\newcommand{\codepath}[1]{\begingroup\ttfamily\small\path{#1}\endgroup}
\DeclareUnicodeCharacter{FF08}{(}
\DeclareUnicodeCharacter{FF09}{)}

\title{NeuroGaze--Distill: Brain-informed Distillation and Depression--Inspired Geometric Priors for Robust Facial Emotion Recognition}

\author{
  Zilin Li\thanks{First author.}%
  \thanks{This work was completed while the author was affiliated with the School of Computer Science and Technology.} \\
  School of Information and Intelligent Science\\
  Donghua University\\
  2999 North Renmin Road 201620 \\
  Shanghai, China\\
  \texttt{tzulamlee@gmail.com} \\
  \And
  Weiwei Xu \\
  School of Information and Intelligent Science\\
  Donghua University\\
  2999 North Renmin Road 201620 \\
  Shanghai, China\\
  \texttt{231310126@mail.dhu.edu.cn} \\
  \And
  Xuanqi Zhao \\
  School of Information and Intelligent Science\\
  Donghua University\\
  2999 North Renmin Road 201620 \\
  Shanghai, China\\
  \texttt{241310629@mail.dhu.edu.cn} \\
  \And
  Yiran Zhu \\
  Department of Computer \\
  North China Electric Power University (BaoDing) \\
  No. 619, Yonghua North Street\\
  Hebei, China\\
  \texttt{ciaran\_study@yeah.net} 
}

\begin{document}
\maketitle
% —— 在这里插入，无编号脚注，不会占用“第一作者”的脚注 1 —— 
\begingroup
\renewcommand\thefootnote{}\footnote{Note on the name. ``NeuroGaze--Distill'' emphasizes neuro-informed distillation. 
Gaze heatmaps are optional and may be disabled in the final experiments; the ``Gaze'' term 
survives to reflect the broader privileged-signal design.}%
\addtocounter{footnote}{-1}%
\endgroup
% —— 到此结束 —— 
% —— 无编号脚注：Preprint —— 
\begingroup
\renewcommand\thefootnote{}\footnote{\textbf{Preprint.} This manuscript is a preprint; it has not been peer reviewed. 
It is shared to facilitate timely dissemination of the research.}%
\addtocounter{footnote}{-1}%
\endgroup

\begin{abstract}
Facial emotion recognition (FER) models trained only on pixels often fail to generalize across datasets because facial appearance is an indirect—and biased—proxy for underlying affect. We present NeuroGaze-Distill, a cross-modal distillation framework that transfers brain-informed priors into an image-only FER student via static Valence–Arousal (V/A) prototypes and a depression-inspired geometric prior (D-Geo). A teacher trained on EEG topographic maps from DREAMER and MAHNOB-HCI produces a consolidated 5×5 V/A prototype grid that is frozen and reused; no EEG–face pairing and no non-visual signals at deployment are required. The student (ResNet-18/50) is trained on FERPlus with conventional CE/KD and two lightweight regularizers: (i) Proto-KD (cosine) aligns student features to the static prototypes; (ii) D-Geo softly shapes the embedding geometry in line with affective findings often reported in depression research (e.g., anhedonia-like contraction in high-valence regions). We evaluate both within-domain (FERPlus validation) and cross-dataset protocols (AffectNet-mini; optional CK+), reporting standard 8-way scores alongside present-only Macro-F1 and balanced accuracy to fairly handle label-set mismatch. Ablations attribute consistent gains to prototypes and D-Geo, and favor 5×5 over denser grids for stability. The method is simple, deployable, and improves robustness without architectural complexity.
\end{abstract}

% ---- Pipeline figure (PDF in project root) ----
\begin{figure}[t]
  \centering
  \includegraphics[width=\linewidth,page=1]{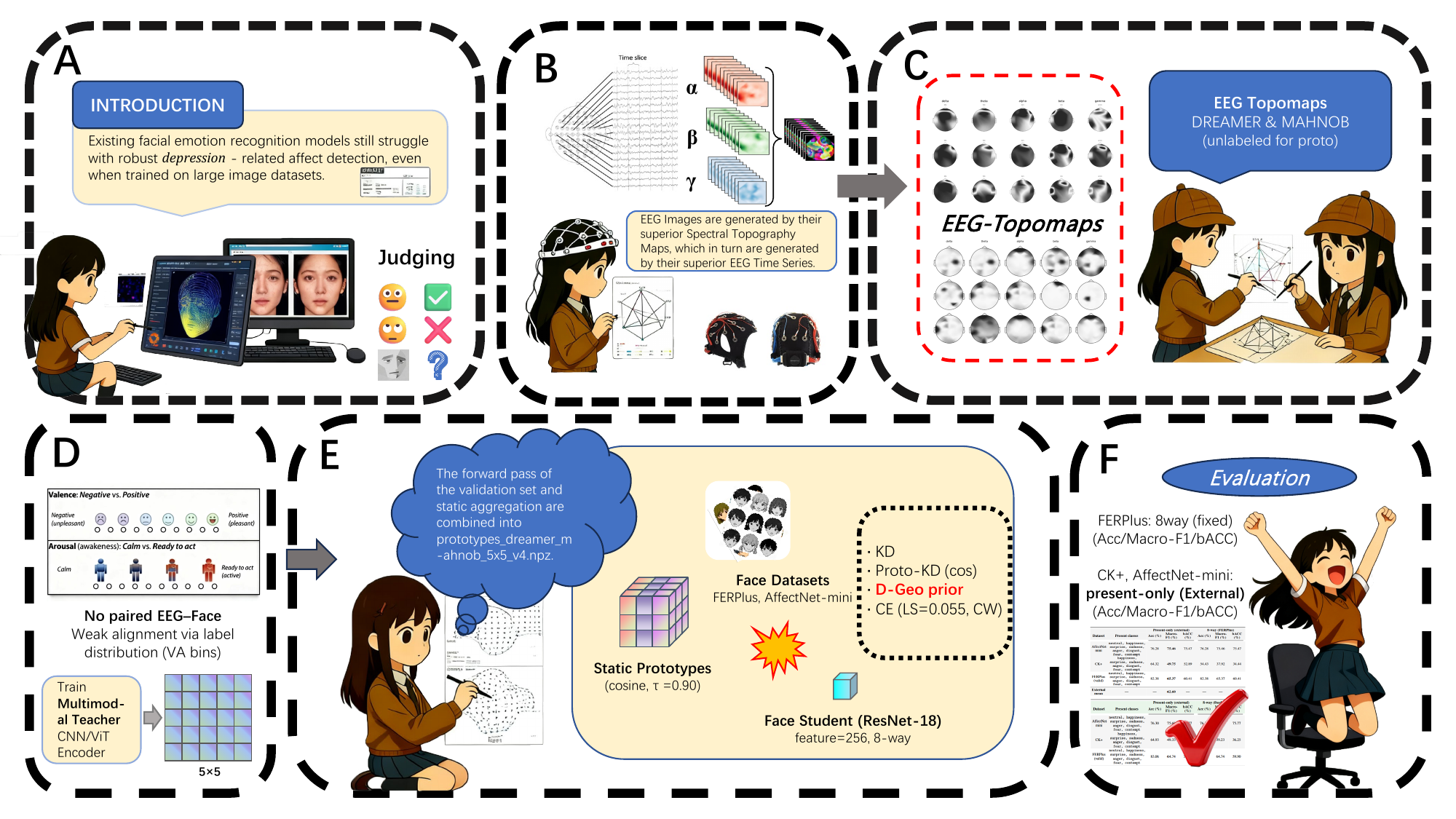}
  \caption{\textbf{NeuroGaze--Distill overview.}
  \textbf{(A)} FER suffers from distribution shift; robust depression–related affect detection is challenging with pixels alone.
  \textbf{(B)} EEG preprocessing: spectral features are rendered as topographic images.
  \textbf{(C)} Teacher data: EEG topomaps from \textsc{DREAMER} and \textsc{MAHNOB--HCI}.
  \textbf{(D)} V/A circumplex with 5$\times$5 binning; no paired EEG--face samples required.
  \textbf{(E)} Distillation: ResNet--18/50 student with CE (LS=0.055, CW), KD on logits, \emph{Proto--KD} (cosine, $\tau\!\approx\!0.90$) toward \emph{static prototypes}, and a light \emph{D--Geo} prior (depression--inspired).
  \textbf{(F)} Evaluation: within--domain on \textsc{FERPlus} and cross--dataset on \textsc{AffectNet--mini} (optional \textsc{CK+}) with \emph{present--only} metrics.
  }
  \label{fig:overview}
\end{figure}
% ---- end pipeline figure ----

\section{Introduction}

% —— 无编号脚注：Artifacts-only reproducibility —— 
\begingroup
\renewcommand\thefootnote{}\footnote{\textbf{Artifacts-only reproducibility.} We do not release full source code at submission time, but we provide a minimal repository to reproduce reported tables/figures: \url{https://github.com/Lixeeone/NeuroGaze-Distill}.}%
\addtocounter{footnote}{-1}%
\endgroup

\noindent\textbf{Motivation.}
Human affect is latent; faces are observable but ambiguous. Cross-dataset shifts in
demographics, capture conditions, and label conventions substantially degrade FER robustness.
In contrast, physiological signals such as EEG encode affective dynamics in a representation
less entangled with appearance. However, collecting paired EEG--face data at scale is
impractical and undesirable for deployable vision-only systems.

\medskip
\noindent\textbf{Idea.}
We learn \emph{static neuro-informed prototypes} in a continuous Valence--Arousal (V/A) space
and distill them into an image-only FER model. A teacher trained on EEG topographic maps
(\textsc{DREAMER}, with \textsc{MAHNOB--HCI} as unlabeled support) regresses V/A; its validation
embeddings are aggregated into a \textbf{5$\times$5} V/A prototype grid, which is then
\emph{frozen} and reused across students. A standard ResNet-18/50 student \cite{he2016deep} trained on
\textsc{FERPlus} \cite{barsoum2016ferplus} receives conventional CE with label smoothing \cite{muller2019labelsmoothing} and
logit KD \cite{hinton2015distilling}, plus two lightweight regularizers:
(i) \emph{Proto--KD} (cosine) to align features with the static prototypes; and
(ii) a \emph{depression-inspired geometric prior (D--Geo)} that
softly shapes the embedding geometry in line with affective findings (e.g., anhedonia) \cite{pizzagalli2014anhedonia,treadway2011reconceptualizing}.
Deployment remains vision-only.

\medskip
\noindent\textbf{Contributions.}
\begin{itemize}
  \item \textbf{Static neuro-informed prototypes.}
        A simple, reusable EEG$\rightarrow$V/A prototype formation requiring neither paired
        EEG--face data nor non-visual inputs at test time, grounded in a V/A circumplex space \cite{russell1980circumplex}.
  \item \textbf{Minimalist loss cocktail.}
        CE (label smoothing \cite{muller2019labelsmoothing} + mild class weights) + logit KD \cite{hinton2015distilling} +
        \textbf{Proto--KD} (cf. prototype learning \cite{snell2017prototypical,movshovitz2017proxynca}) + \textbf{D--Geo},
        improving cross-dataset generalization without architectural complexity.
  \item \textbf{Depression-aware perspective.}
        A non-diagnostic geometric prior that regularizes representation structure using insights
        from affective neuroscience \cite{pizzagalli2014anhedonia,treadway2011reconceptualizing}.
\end{itemize}

\section{Background and Related Work}
\label{sec:background}

\begin{figure}[H]
  \centering
  \newcommand{\imgH}{5.2cm}
  \begin{minipage}[t]{0.47\linewidth}
    \centering
    \includegraphics[height=\imgH]{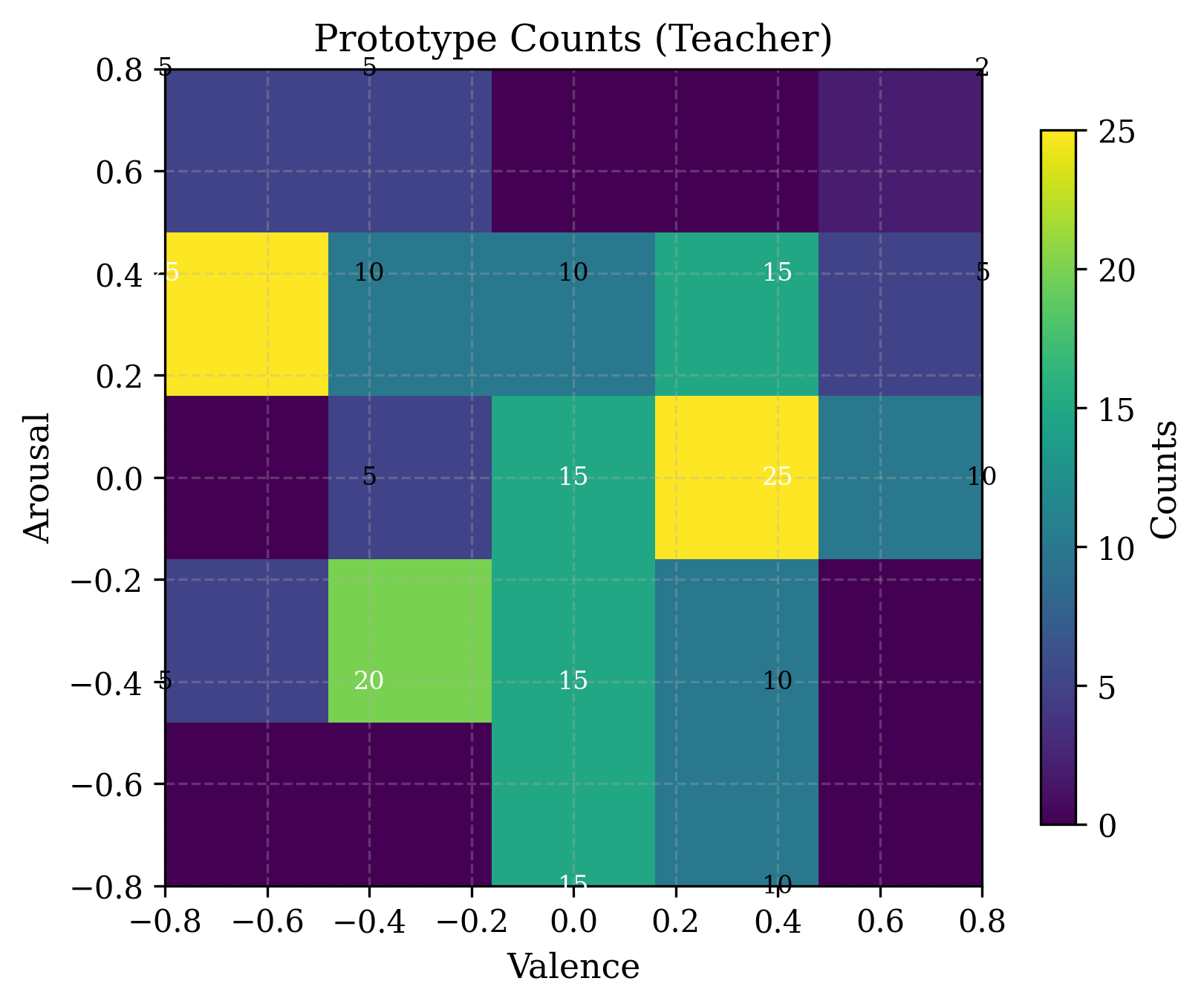}
  \end{minipage}\hfill
  \begin{minipage}[t]{0.51\linewidth}
    \centering
    \includegraphics[height=\imgH]{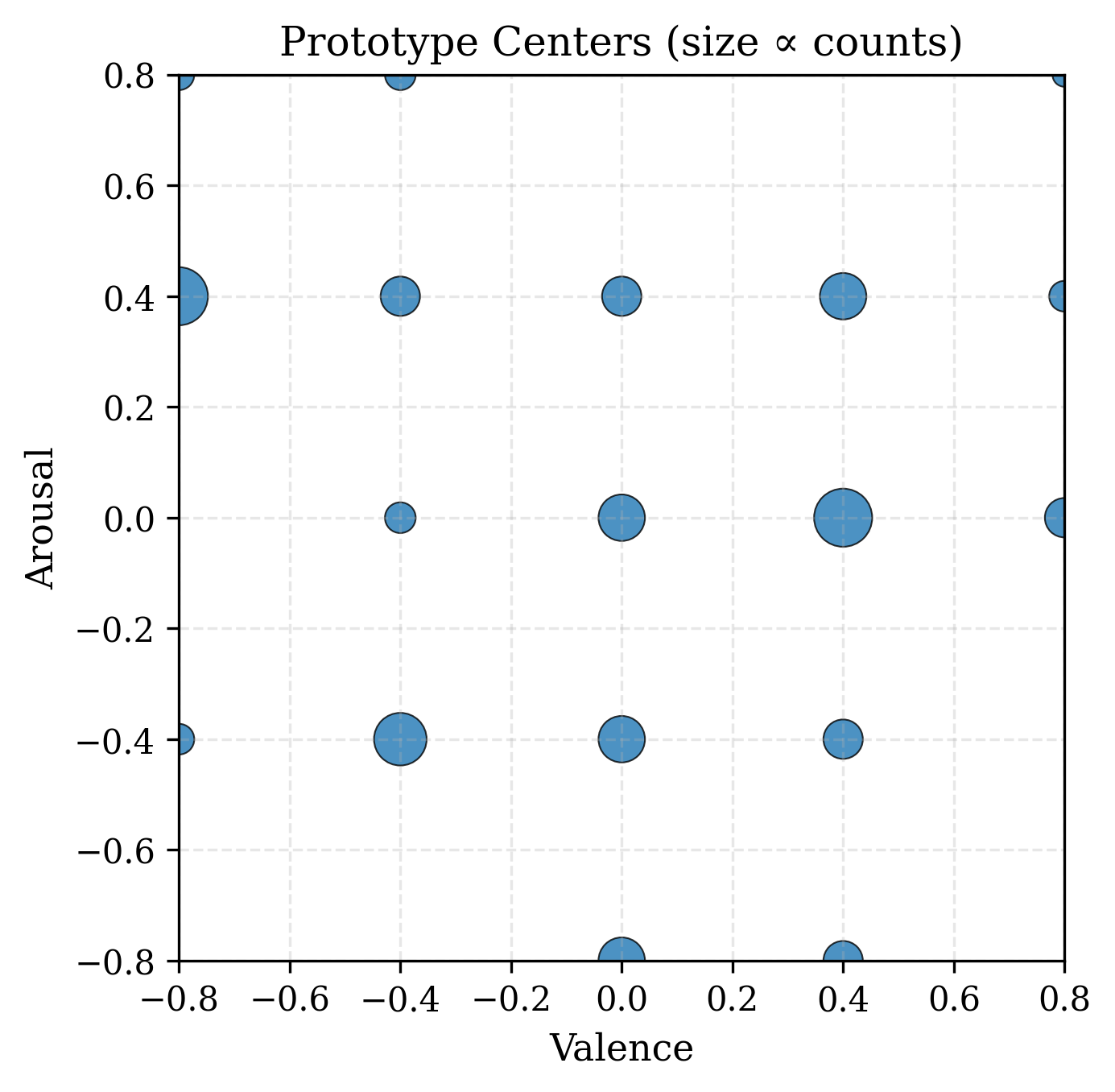}
  \end{minipage}
  \caption{\textbf{Prototype coverage ($5\times 5$, v4).}
  \textbf{Left:} counts per V/A bin; \textbf{Right:} grid centers with marker size $\propto$ counts.
  Panels are height-matched and width-balanced (no cropping).}
  \label{fig:proto_grid}
\end{figure}

\paragraph{Affective spaces.}
Following the circumplex view of affect \cite{russell1980circumplex},
we adopt a continuous V/A space and discretize it for learning.
A fixed \textbf{$5\times 5$} grid balances coverage and statistical stability: denser grids (e.g., $7\times 7$) increase sparsity and bin collapse in underrepresented regions, while coarser grids lose resolution near decision boundaries.
Fig.~\ref{fig:proto_grid} (left) visualizes the per-bin coverage of our teacher-derived data; Fig.~\ref{fig:proto_grid} (right) shows the grid centers with marker sizes proportional to counts.
This V/A geometry serves as the \emph{teacher space} for forming static prototypes, while the student remains a categorical FER classifier, decoupling continuous affect structure from the final deployment task.

\paragraph{FER and distribution shift.}
Deep FER has improved in-domain accuracy, yet cross-dataset robustness remains fragile due to differences in demographics, capture devices, annotation protocols, and class prevalence.
To avoid confounding from missing classes on the target set, we report both the standard \emph{8-way} metrics and \emph{present-only} metrics that restrict evaluation to labels actually present in the target dataset, providing a fairer measure of transferability.

\paragraph{Physiological signals for affect.}
Physiological channels (EEG, EDA, HRV, gaze) capture affective dynamics with noise and bias characteristics different from pixels.
EEG in particular provides time--frequency measurements from distributed sensors; topographic images (``topomaps'') can be rendered by interpolating per-channel band power onto a 2-D scalp layout.
Most prior works either perform EEG-only recognition or require paired multimodal training.
In contrast, we use EEG only to \emph{form} a compact prior---static V/A prototypes---and then train a vision-only student without any paired EEG--face examples or non-visual inputs at deployment, using publicly available datasets such as DREAMER \cite{katsigiannis2018dreamer} and MAHNOB--HCI \cite{soleymani2012mahnob}.

\paragraph{Depression-informed priors.}
Affective and clinical literature commonly discusses altered reward processing and \emph{anhedonia} in depressive conditions \cite{pizzagalli2014anhedonia,treadway2011reconceptualizing}.
We operationalize this insight as a light \textbf{D--Geo} prior: encourage controlled compactness for features associated with high-valence regions while maintaining separability (margins) elsewhere.
The prior is non-diagnostic, applied uniformly across the dataset, and interacts additively with standard objectives.

\paragraph{Knowledge distillation and prototypes.}
Knowledge distillation (KD) transfers information from a teacher to a student via logit-based soft targets \cite{hinton2015distilling} (and early model compression \cite{bucilua2006model}), feature alignment, or relation matching.
Prototype-based learning summarizes class/region structure with representative vectors \cite{snell2017prototypical,movshovitz2017proxynca}.
Our framework combines both: the student receives CE+KD while also aligning, via a cosine objective, to a fixed bank of \emph{EEG-derived V/A prototypes}.

% -------------------- Figure 2 (scaled, no trim) --------------------
% ------------------ end Figure 2 ------------------

\section{Datasets and Preprocessing}
\label{sec:data}

\subsection{EEG teacher data}
\textbf{DREAMER.}
We use the public DREAMER affect dataset \cite{katsigiannis2018dreamer} and render EEG \emph{topographic images} (``topomaps'')
from band power ($\delta,\theta,\alpha,\beta,\gamma$) computed on stimulus segments.
Per subject and per band we z-score and min–max normalize to $[0,1]$ for visualization consistency;
maps are exported at a fixed resolution for training the EEG teacher.

\textbf{MAHNOB--HCI.}
We follow the same pipeline for MAHNOB--HCI \cite{soleymani2012mahnob}. Where gaze is available, we may export a heatmap
for analysis only; no non-visual signals are used by the student at training or deployment.
For reproducibility we maintain a manifest (\texttt{mahnob\_topomaps\_manifest.csv}) and a consolidated archive
(\texttt{mahnob\_topomaps\_all.npz}) to reproduce the teacher’s validation features used for prototype formation.
Illustrative topomap grids are shown in Fig.~\ref{fig:topomap_grids}.

\subsection{Face student data}
\textbf{FERPlus.}
We use pre-packed NPZs with $\![N,48,48]\!$ grayscale images and 8-class label distributions
(\texttt{ferplus\_train/valid/test.npz}), together with manifests
(\texttt{ferplus\_manifest\_*.csv}) and a class map (\texttt{class\_map\_ferplus.json}).
Images are center-aligned, mean–std normalized, and augmented with safe transforms
(random crop/flip and mild color jitter) that do not target identity cues.
\emph{No face exemplars are displayed in this paper; we report only aggregate metrics and non-identifiable visualizations.}
We follow the protocol of FERPlus \cite{barsoum2016ferplus}.

\textbf{AffectNet-mini.}
We adopt a reduced AffectNet split with CSV labels
(\texttt{labels\_train/valid/test.csv}, \texttt{labels\_all.csv}) and \texttt{class\_map.json}, based on AffectNet \cite{mollahosseini2019affectnet}.
For cross-dataset transfer we report both standard 8-way metrics and \emph{present-only} metrics
restricted to classes available in the target split (Sec.~\ref{sec:background}).
Optionally, we also evaluate on CK+ \cite{lucey2010ckplus}.

\subsection{Processing and binning}
\textbf{EEG $\rightarrow$ V/A.}
Teacher networks regress Valence–Arousal (V/A) from topomaps.
We linearly map reported V/A to $[-1,1]$, then discretize the continuous space with a fixed
\textbf{$5\times 5$} grid (centers from $-0.8$ to $0.8$ on each axis).
Validation embeddings are aggregated per bin to form 25 static prototypes.

\textbf{Faces (categorical) $\rightarrow$ student.}
The student remains an 8-way FER classifier (ResNet--18/50, 256-D feature).
Training uses CE with label smoothing \cite{muller2019labelsmoothing} and mild class weights, logit KD at $T{=}5.0$ \cite{hinton2015distilling},
Proto--KD (cosine) toward the static prototype bank, and a light depression-inspired geometric prior (D--Geo) \cite{pizzagalli2014anhedonia,treadway2011reconceptualizing}.
All figures and tables are exported automatically to \texttt{viz/} and \texttt{outs/}.

% -------------------- EEG topomap grids (shifted left as a block) --------------------
\begin{figure}[H]
  \centering
  \hspace*{-0.025\linewidth}
  \newcommand{\imgH}{4.6cm}
  \begin{minipage}[t]{0.49\linewidth}
    \centering
    \includegraphics[height=\imgH,page=1]{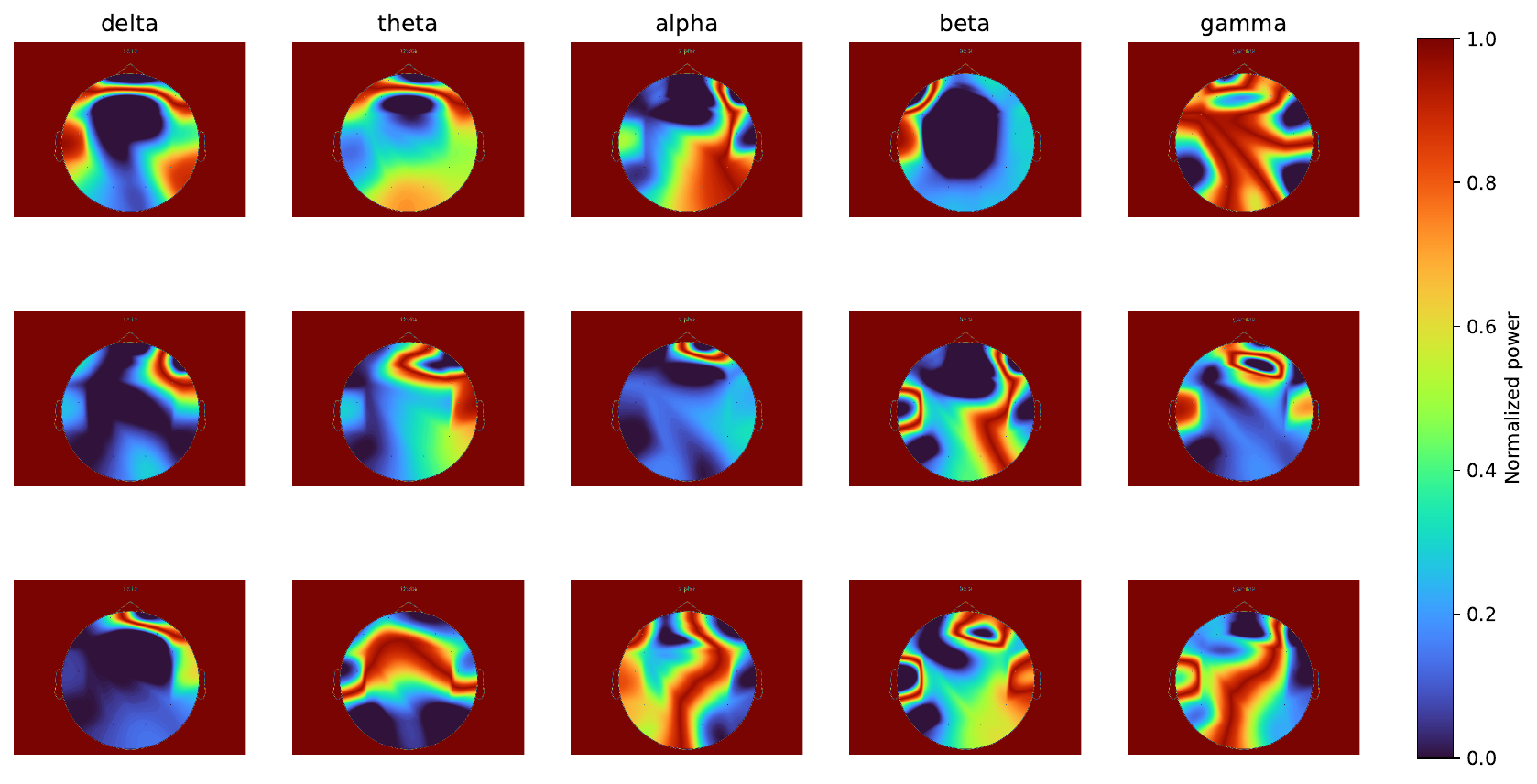}
  \end{minipage}\hfill
  \begin{minipage}[t]{0.45\linewidth}
    \centering
    \includegraphics[height=\imgH,page=1]{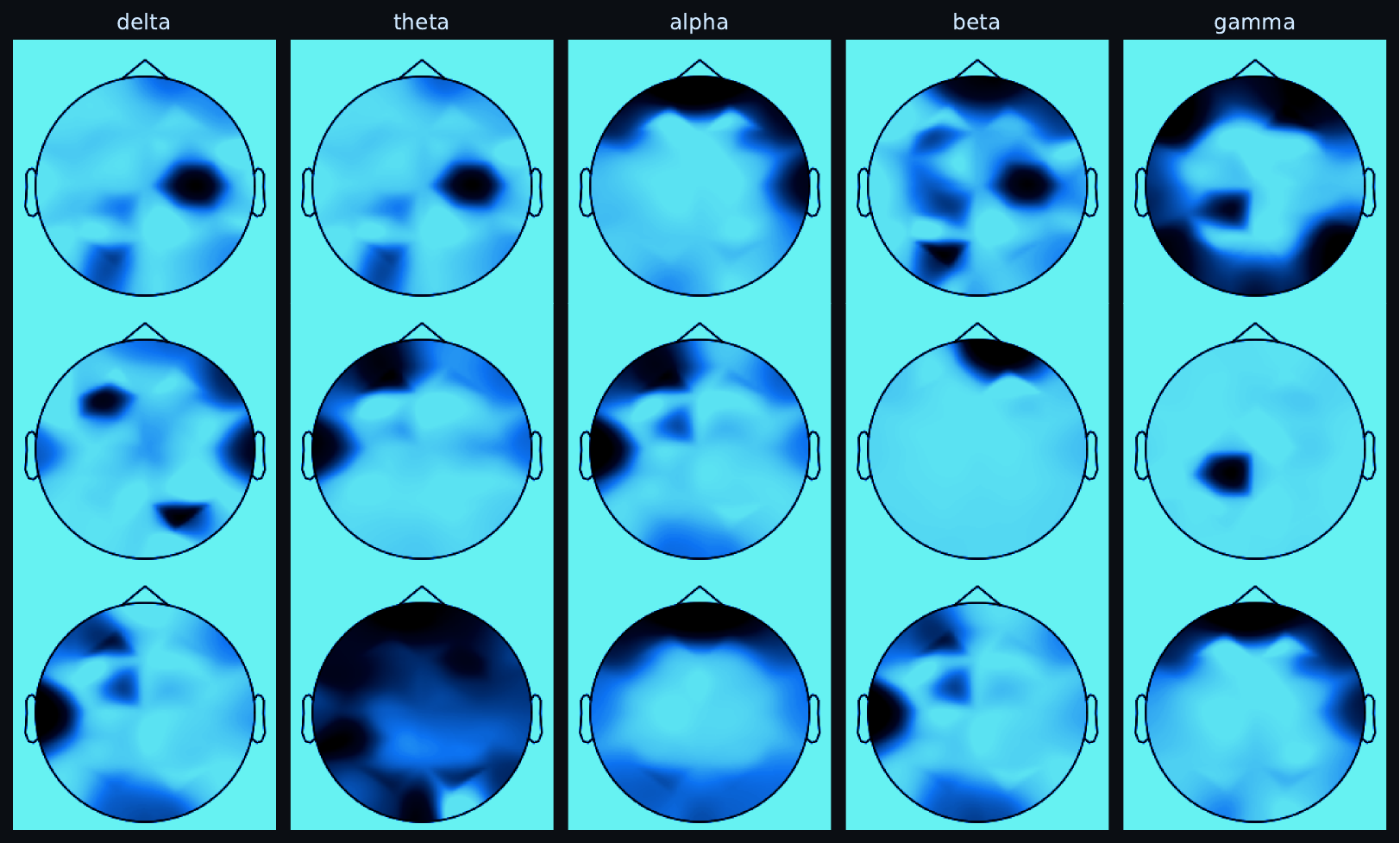}
  \end{minipage}
  \caption{\textbf{EEG topomap grids used by the teacher (Sec.~\ref{sec:data}).}
\textbf{Left:} DREAMER band-power topomaps \cite{katsigiannis2018dreamer}.
\textbf{Right:} MAHNOB--HCI in a cool-blue palette \cite{soleymani2012mahnob}.
Maps are per-band normalized for visualization and rendered with 10--20 style interpolation; they are non-identifiable.
These grids train the EEG teacher that regresses V/A, after which validation embeddings are aggregated into a fixed
$5{\times}5$ V/A prototype bank (Sec.~\ref{sec:data}, Processing and binning).
No gaze or other non-visual signals are used by the student.}
  \label{fig:topomap_grids}
\end{figure}
% ------------------------------------------------------------------------------- %

\section{Method}
\label{sec:method}

\subsection{Teacher and static prototype formation}
\label{sec:teacher}
We train a CNN/ViT teacher \cite{he2016deep,dosovitskiy2021vit} to regress Valence--Arousal (V/A) from EEG topographic maps (Sec.~\ref{sec:data}).
On the teacher \emph{validation} split, we map continuous V/A into a fixed $5{\times}5$ grid (centers in $[-0.8,0.8]$), and average
the penultimate features within each bin to obtain 25 static prototypes $\mathcal{P}{=}\{p_k\}_{k=1}^{25}$.
This yields the consolidated \textbf{v4} prototype bank (\texttt{prototypes\_dreamer\_mahnob\_5x5\_v4.npz}), formed on
DREAMER with unlabeled MAHNOB--HCI as a stability supplement; the bank is \emph{frozen} and reused across all students and datasets
(no paired EEG--face samples are required, and no non-visual signals are used at deployment).
Fig.~\ref{fig:proto_grid} visualizes prototype coverage, motivating the $5{\times}5$ choice over denser grids for robustness.

\subsection{Student network}
\label{sec:student}
The student is a standard ResNet--18/50 backbone \cite{he2016deep} with a 256-D projection and an 8-way classifier. We adopt
channels-last memory format, mixed precision (AMP), gradient clipping, and label smoothing ($\alpha{=}0.055$) \cite{muller2019labelsmoothing}.
Unless otherwise stated, \emph{student EMA is disabled} (Mean-Teacher style EMA \cite{tarvainen2017meanteacher} underperforms here), and we do not employ additional LDACC-like losses.

\subsection{Losses}
\label{sec:losses}
Let $x$ be an input face, $f(x)\in\mathbb{R}^{256}$ the L2-normalized feature, $z(x)\in\mathbb{R}^{8}$ the logits,
$y$ the (smoothed) 8-class target distribution, and $\mathcal{P}$ the fixed prototype set.

\paragraph{Cross-entropy (CE).}
We use CE with label smoothing ($\alpha{=}0.055$) \cite{muller2019labelsmoothing} and mild class weights to stabilize training on imbalanced emotions.

\paragraph{Logit distillation (KD).}
We match softened student logits to a vision teacher with temperature $T{=}5.0$ using an MSE/KL objective with a medium loss weight
\cite{hinton2015distilling}.

\paragraph{Prototype distillation (Proto--KD, cosine).}
For each sample, we compute cosine similarities $s_k{=}\cos\!\big(f(x),p_k\big)$ to all prototypes and obtain a soft bin distribution
$q^{\text{stu}}{=}\operatorname{softmax}(s/\tau)$ with feature temperature $\tau{=}0.90$.
The prototype prior $q^{\text{pro}}$ is the (frozen) per-bin prior induced by $\mathcal{P}$.
We minimize $D_{\mathrm{KL}}\!\left(q^{\text{pro}}\Vert q^{\text{stu}}\right)$ with a small weight (cf. prototype learning \cite{snell2017prototypical,movshovitz2017proxynca}).

\paragraph{Depression-inspired geometric prior (D--Geo).}
D--Geo is a weak, \emph{non-diagnostic} regularizer on representation geometry. Concretely,
(i) for high-valence categories (we use \{happiness, surprise\}), we apply a light within-class variance cap;
(ii) globally, we encourage inter-class margins to preserve separability. The term is \emph{late-activated} with a cosine ramp
(epochs 20$\rightarrow$60) and a small weight, motivated by anhedonia-related findings \cite{pizzagalli2014anhedonia,treadway2011reconceptualizing}.

\paragraph{Overall objective.}
\[
\mathcal{L} \;=\; \underbrace{\mathcal{L}_{\mathrm{CE}}}_{\text{CE (LS+CW)}} \;+\;
\lambda_{\mathrm{kd}}\,\underbrace{\mathcal{L}_{\mathrm{KD}}}_{\text{MSE/KL, }T{=}5.0}\;+\;
\lambda_{\mathrm{proto}}\,\underbrace{D_{\mathrm{KL}}\!\left(q^{\mathrm{pro}}\Vert q^{\mathrm{stu}}\right)}_{\text{Proto--KD (cos), }\tau{=}0.90}\;+\;
\lambda_{\mathrm{geo}}\,\underbrace{\mathcal{L}_{\mathrm{D\text{-}Geo}}}_{\text{late-activated}}.
\]
We keep all terms small and stable; together they improve cross-dataset generalization without architectural complexity.

% ===================== Algorithm Box 1 =====================
\vspace{-2pt}
\begin{algorithm}[H]
\caption{Static prototype formation from EEG topomaps (v4 bank)}
\label{alg:proto_formation}
\begin{algorithmic}[1]
\Require Teacher $T$ (CNN/ViT); validation set $\mathcal{D}_{\mathrm{val}}=\{(M_i,\mathbf{v}_i)\}$ with topomaps $M_i$ and V/A $\mathbf{v}_i\in[-1,1]^2$; grid size $G{=}5$; centers $\mathcal{C}{=}\{-0.8,\ldots,0.8\}$
\Ensure Frozen prototype bank $\mathcal{P}{=}\{p_k\}_{k=1}^{K}$, $K{=}G^2$; per-bin prior $q^{\mathrm{pro}}\in\mathbb{R}^K$
\State Initialize counts $N_{u,v}\!\leftarrow\!0$, sums $S_{u,v}\!\leftarrow\!\mathbf{0}$ for all $(u,v)\in\{1,\dots,G\}^2$
\ForAll{$(M_i,\mathbf{v}_i)\in\mathcal{D}_{\mathrm{val}}$} \Comment{No EEG--face pairing is required}
  \State $\mathbf{e}_i \leftarrow \text{L2Norm}\!\big(\textsc{Penultimate}(T(M_i))\big)$
  \State $(u,v) \leftarrow \textsc{BinVA}(\mathbf{v}_i;\mathcal{C})$ \Comment{map V/A to $G{\times}G$}
  \State $S_{u,v} \mathrel{+}= \mathbf{e}_i$; \quad $N_{u,v} \mathrel{+}= 1$
\EndFor
\ForAll{$(u,v)$}
  \If{$N_{u,v}>0$}
     \State $p_{u,v}\leftarrow S_{u,v}/N_{u,v}$
  \Else
     \State $p_{u,v}\leftarrow \textsc{NearestNonEmptyMean}(S,N)$ \Comment{fill empty bin by nearest bin mean}
  \EndIf
\EndFor
\State $\mathcal{P}\leftarrow \{p_{u,v}\}_{u,v}$; \quad
       $q^{\mathrm{pro}}_{u,v} \leftarrow \dfrac{N_{u,v}+\varepsilon}{\sum_{a,b}(N_{a,b}+\varepsilon)}$ \Comment{Laplace smoothing $\varepsilon{\approx}1$}
\State \textsc{SaveNPZ}(\texttt{prototypes\_dreamer\_mahnob\_5x5\_v4.npz}; $\mathcal{P},q^{\mathrm{pro}}$)
\State \Return $\mathcal{P},\,q^{\mathrm{pro}}$
\vspace{2pt}
\Statex \hrulefill
\Function{BinVA}{$\mathbf{v};\mathcal{C}$}
  \State $u\!\leftarrow\!\argmin_j\,|\mathbf{v}_{\text{val}}-\mathcal{C}_j|$; \quad
         $v\!\leftarrow\!\argmin_j\,|\mathbf{v}_{\text{aro}}-\mathcal{C}_j|$; \Return $(u,v)$
\EndFunction
\Function{NearestNonEmptyMean}{$S,N$}
  \State \Return mean of nearest $(a,b)$ with $N_{a,b}{>}0$ (fallback: global mean)
\EndFunction
\end{algorithmic}
\end{algorithm}
\vspace{-4pt}
% =================== End Algorithm Box 1 ===================

% ===================== Algorithm Box 2 =====================
\vspace{-2pt}
\begin{algorithm}[H]
\caption{Student training with CE + KD + Proto--KD + D--Geo}
\label{alg:student_training}
\begin{algorithmic}[1]
\Require Student $F{=}\textsc{Head}\circ\textsc{Proj}\circ\textsc{Backbone}$; optional vision teacher $C$ (for KD);
prototypes $\mathcal{P}$ and prior $q^{\mathrm{pro}}$; label smoothing $\alpha{=}0.055$;
KD temperature $T{=}5.0$; Proto--KD temperature $\tau{=}0.90$;
weights $(\lambda_{\mathrm{kd}},\lambda_{\mathrm{proto}},\lambda_{\mathrm{geo}})$;
late schedule $s_{\mathrm{geo}}(e)$ active on epochs $20{\rightarrow}60$; class weights $w_c$
\Ensure Trained parameters $\theta_F$
\For{$e \gets 1$ \textbf{to} $E$} \Comment{AMP + channels-last + grad-clip(1.0) applied}
  \For{minibatch $\{(x_b,y_b)\}_{b=1}^{B}$}
    \State $\mathbf{f}_b \gets \mathrm{L2Norm}\!\big(\textsc{Proj}(\textsc{Backbone}(x_b))\big)\in\mathbb{R}^{256}$;\quad
           $\mathbf{z}_b \gets \textsc{Head}(\mathbf{f}_b)\in\mathbb{R}^{8}$
    \State \textbf{CE:} $\tilde{y}_b \gets \textsc{LabelSmooth}(y_b;\alpha)$;\quad
           $\mathcal{L}_{\mathrm{ce}} \gets \sum_b \mathrm{CE}(\tilde{y}_b,\mathbf{z}_b; w_c)$
    \If{$C$ exists}
      \State $\mathbf{z}^T_b \gets C(x_b)$
      \State $\mathcal{L}_{\mathrm{kd}} \gets \sum_b \mathrm{KL}\!\big(\mathrm{Softmax}(\mathbf{z}^T_b/T)\,\Vert\,\mathrm{Softmax}(\mathbf{z}_b/T)\big)$
    \Else
      \State $\mathcal{L}_{\mathrm{kd}} \gets 0$
    \EndIf
    \State \textbf{Proto--KD:}
      $\mathbf{s}_b \gets [\cos(\mathbf{f}_b,p_k)]_{k=1}^{K}$;\quad
      $q^{\mathrm{stu}}_b \gets \mathrm{Softmax}(\mathbf{s}_b/\tau)$;\quad
      $\mathcal{L}_{\mathrm{proto}} \gets \sum_b \mathrm{KL}\!\big(q^{\mathrm{pro}}\,\Vert\,q^{\mathrm{stu}}_b\big)$
    \State \textbf{D--Geo (late):} compute per-class means $\mu_c$ and variances $\sigma^2_c$ on the minibatch;
           let $\mathcal{H}{=}\{\texttt{happiness},\texttt{surprise}\}$
    \State $\mathcal{L}_{\mathrm{var}} \gets \sum_{c\in\mathcal{H}} \max\!\big(0,\ \sigma^2_c - \sigma^2_{\max}\big)$;\quad
           $\mathcal{L}_{\mathrm{margin}} \gets \sum_{c\neq c'} \max\!\big(0,\ m - \lVert \mu_c-\mu_{c'}\rVert_2\big)$
    \State $\mathcal{L}_{\mathrm{geo}} \gets s_{\mathrm{geo}}(e)\,\big(\alpha_{\mathrm{var}}\mathcal{L}_{\mathrm{var}} + \alpha_{\mathrm{mar}}\mathcal{L}_{\mathrm{margin}}\big)$
    \State \textbf{Total:} $\mathcal{L} \gets \mathcal{L}_{\mathrm{ce}}
           + \lambda_{\mathrm{kd}}\mathcal{L}_{\mathrm{kd}}
           + \lambda_{\mathrm{proto}}\mathcal{L}_{\mathrm{proto}}
           + \lambda_{\mathrm{geo}}\mathcal{L}_{\mathrm{geo}}$
    \State \textsc{Backprop}$(\mathcal{L});\ \textsc{ClipGrad}(1.0);\ \textsc{Step}(\text{AdamW}, \text{cosine LR})$
  \EndFor
\EndFor
\State \Return $\theta_F$
\end{algorithmic}
\end{algorithm}
\vspace{-4pt}
% =================== End Algorithm Box 2 ===================

\subsection{Implementation details}
\label{sec:impl}
Unless otherwise stated, \emph{all results in Sec.~\ref{sec:exp} use the same training recipe.}
We train with AdamW \cite{loshchilov2019adamw} (Adam \cite{kingma2015adam} variant) and a cosine schedule \cite{loshchilov2017sgdr};
base learning rate $2\times10^{-4}$, weight decay $0.05$,
batch size $128$, mixed precision (AMP), channels-last, and gradient clipping $(1.0)$.
Random seeds are fixed, logs are recorded with TensorBoard, and figure/table/CSV exporters
write to \texttt{viz/} and \texttt{outs/}. All released artifacts (the static prototype bank
and student checkpoints) are versioned with SHA-256; filenames and digests are summarized in
Table~\ref{tab:sha256}.

\paragraph{Model variants compared in Sec.~5.2.}
All students share the backbone and the above training recipe; they differ \emph{only} in loss terms:
\mbox{B0}~CE only; \mbox{B1}~CE + logit KD ($T{=}5.0$); \mbox{B2}~B1 + Proto--KD (cosine, $\tau{=}0.90$);
\mbox{B3}~B2 + D--Geo (full method, late activation with a small weight).
A gaze-augmented EEG teacher can be used when available, but in our final runs gaze is disabled for
consistency across datasets.

\begin{table}[H]
  \centering
  \small
  \setlength{\tabcolsep}{6pt}
  \newcolumntype{Y}{>{\centering\arraybackslash}X}
  \begin{tabularx}{\linewidth}{@{}l Y Y@{}}
    \toprule
    \textbf{Artifact} & \textbf{Filename (in supplementary archive)} & \textbf{SHA-256} \\
    \midrule
    Prototypes (v4, 5$\times$5)
      & \url{artifacts/prototypes_dreamer_mahnob_5x5_v4.npz}
      & \seqsplit{bc6ea0c9aa3ef4415772996bdde2d3f5945f1242d65a3c9e2096a993efcc7432} \\
    Student ckpt (A3\_full, 100ep)
      & \url{artifacts/student_A3_full_100.ckpt}
      & \seqsplit{5647a905d7f4b3600abb32e80d1abd04e0b326222b273bfab488c695751a82f5} \\
    Metrics bundle (FERPlus valid)
      & \url{artifacts/metrics_ferplus_valid_abla_A3_full.json}
      & \seqsplit{95861ecebd39b988bd549916393cffcc3b1b0d7cbc2fe0ae2f79662789c0b217} \\
    \bottomrule
  \end{tabularx}
  \caption{Artifacts provided privately to reviewers. Filenames are relative to the supplementary archive; SHA-256 digests enable verification.}
  \label{tab:sha256}
\end{table}

\section{Experiments}
\label{sec:exp}

\subsection{Protocols and metrics}
\label{sec:protocols}
We train on FERPlus and evaluate both within-domain and under cross-dataset shift.
\textbf{Within-domain} results are reported on the FERPlus validation split with Accuracy (Acc), Macro-F1,
and balanced accuracy (bACC; mean of per-class recall) \cite{brodersen2010balanced}. Macro-AUROC can be added for completeness.
Unless otherwise noted, we report the mean over 3 seeds and 95\% confidence intervals
via stratified bootstrapping (1{,}000 resamples) \cite{efron1994bootstrap}.
\textbf{Cross-dataset} evaluation applies the FERPlus-trained student to AffectNet-mini (and optionally CK+).
Following our label-mismatch discussion, we report both: (i) \emph{present-only} metrics computed using only
the classes available in the target set, and (ii) the full \emph{8-way} mapping (fixed FER taxonomy).

\subsection{Baselines and variants}
\label{sec:baselines}
We compare four progressively augmented students (same backbone and training recipe; Sec.~\ref{sec:impl}):
\textbf{B0} CE only; \textbf{B1} B0 + logit KD ($T{=}5.0$) \cite{hinton2015distilling}; \textbf{B2} B1 + Proto--KD (cosine; $\tau{=}0.90$;
cf.~\cite{snell2017prototypical,movshovitz2017proxynca});
\textbf{B3} B2 + D--Geo (full method; late activation with a small weight, motivated by \cite{pizzagalli2014anhedonia,treadway2011reconceptualizing}).
A gaze-augmented EEG teacher can be used when available, but in our final runs gaze is disabled
for consistency across datasets.

% ===================== Cross-dataset main table =====================
\begin{table}[H]
\centering
\footnotesize
\setlength{\tabcolsep}{3pt}
\begin{tabular}{>{\centering\arraybackslash}m{1.2cm} >{\centering\arraybackslash}m{3.2cm} *{6}{>{\centering\arraybackslash}m{1.2cm}}}
\toprule
\rowcolor{headgreen} &  & \multicolumn{3}{c}{\textbf{Present-only (external)}} & \multicolumn{3}{c}{\textbf{8-way (fixed mapping)}} \\
\rowcolor{headgreen}\textbf{Dataset} & \textbf{Present classes} & \textbf{Acc (\%)} & \textbf{Macro-F1 (\%)} & \textbf{bACC (\%)} & \textbf{Acc (\%)} & \textbf{Macro-F1 (\%)} & \textbf{bACC (\%)} \\
\cmidrule(lr){3-5} \cmidrule(lr){6-8}
\rowcolor{extrow}
AffectNet-mini & \ttfamily neutral, happiness, surprise, sadness, anger, disgust, fear, contempt & 76.30 & 75.60 & 75.77 & 76.30 & 75.60 & 75.77 \\
\rowcolor{extrow}
CK+ \cite{lucey2010ckplus} & \ttfamily happiness, surprise, sadness, anger, disgust, fear, contempt & 64.93 & 49.33 & 52.46 & 55.86 & 39.23 & 36.25 \\
FERPlus (valid) \cite{barsoum2016ferplus} & \ttfamily neutral, happiness, surprise, sadness, anger, disgust, fear, contempt & 83.06 & 64.74 & 59.90 & 83.06 & 64.74 & 59.90 \\
\bottomrule
\end{tabular}
\caption{Cross-dataset evaluation of \textbf{A3\_full} (D--Geo enabled, 100 epochs). Present-only uses labels present in each target dataset; 8-way uses a fixed FER mapping.}
\label{tab:xval_main}
\end{table}
% ================================================================================

% ===================== Ablation table =====================
\begin{table}[H]
\centering
\footnotesize
\setlength{\tabcolsep}{3pt}
\begin{tabular}{>{\centering\arraybackslash}m{3cm} *{3}{>{\centering\arraybackslash}m{1.8cm}}}
\toprule
\rowcolor{headgreen}\textbf{Variant} & \textbf{Acc (\%)} & \textbf{Macro-F1 (\%)} & \textbf{bACC (\%)} \\
\midrule
CE only & 78.22 & 51.29 & 49.77 \\
+KD \cite{hinton2015distilling} & 82.31 & 63.56 & 59.28 \\
+KD+Proto \cite{snell2017prototypical,movshovitz2017proxynca} & 81.48 & 64.21 & 60.30 \\
Full (+D-Geo) \cite{pizzagalli2014anhedonia,treadway2011reconceptualizing} & 83.06 & 64.74 & 59.90 \\
Full (T=1) & 83.66 & 65.39 & 61.32 \\
\bottomrule
\end{tabular}
\caption{FERPlus validation ablation (8-way). B0$\rightarrow$B3 corresponds to CE $\rightarrow$ +KD $\rightarrow$ +Proto--KD $\rightarrow$ +D--Geo.}
\label{tab:abla_lite}
\end{table}
% ======================================================================

\paragraph{At-a-glance.}
From Table~\ref{tab:abla_lite}, adding KD to CE (B0$\rightarrow$B1) yields a large gain in Macro-F1,
and Proto--KD further improves class balance (bACC). Introducing D--Geo (B3) preserves those gains while
nudging high-valence structure; it gives the best or second-best Macro-F1 on FERPlus valid.
In cross-dataset tests (Table~\ref{tab:xval_main}), \emph{present-only} is consistently higher than \emph{8-way} on CK+,
avoiding penalties for absent classes; on AffectNet-mini the gap is small due to taxonomy alignment.

\subsection{Ablations}
\label{sec:ablations}
We ablate the components on \textsc{FERPlus} (valid). Unless noted, the backbone and recipe are fixed (Sec.~\ref{sec:student}, \ref{sec:losses}).

\paragraph{Proto--KD weight.}
Sweeping $\lambda_{\text{proto}}\!\in\!\{0,0.10,0.12,0.15\}$ shows a stable optimum around $0.12$.

\paragraph{D--Geo schedule/weight.}
A \emph{late} cosine ramp (epochs $20{\rightarrow}60$) with a small weight performs best; enabling from epoch~0 slightly harms early separability.

\paragraph{Grid size.}
A $5{\times}5$ V/A grid outperforms $7{\times}7$ by avoiding sparse/collapsing bins (see Fig.~\ref{fig:topomap_grids}).

\paragraph{EMA.}
Teacher EMA helps prototype stability; student EMA underperforms here and is disabled \cite{tarvainen2017meanteacher}.

\begin{figure}[H]
  \centering
  \includegraphics[width=0.7\linewidth]{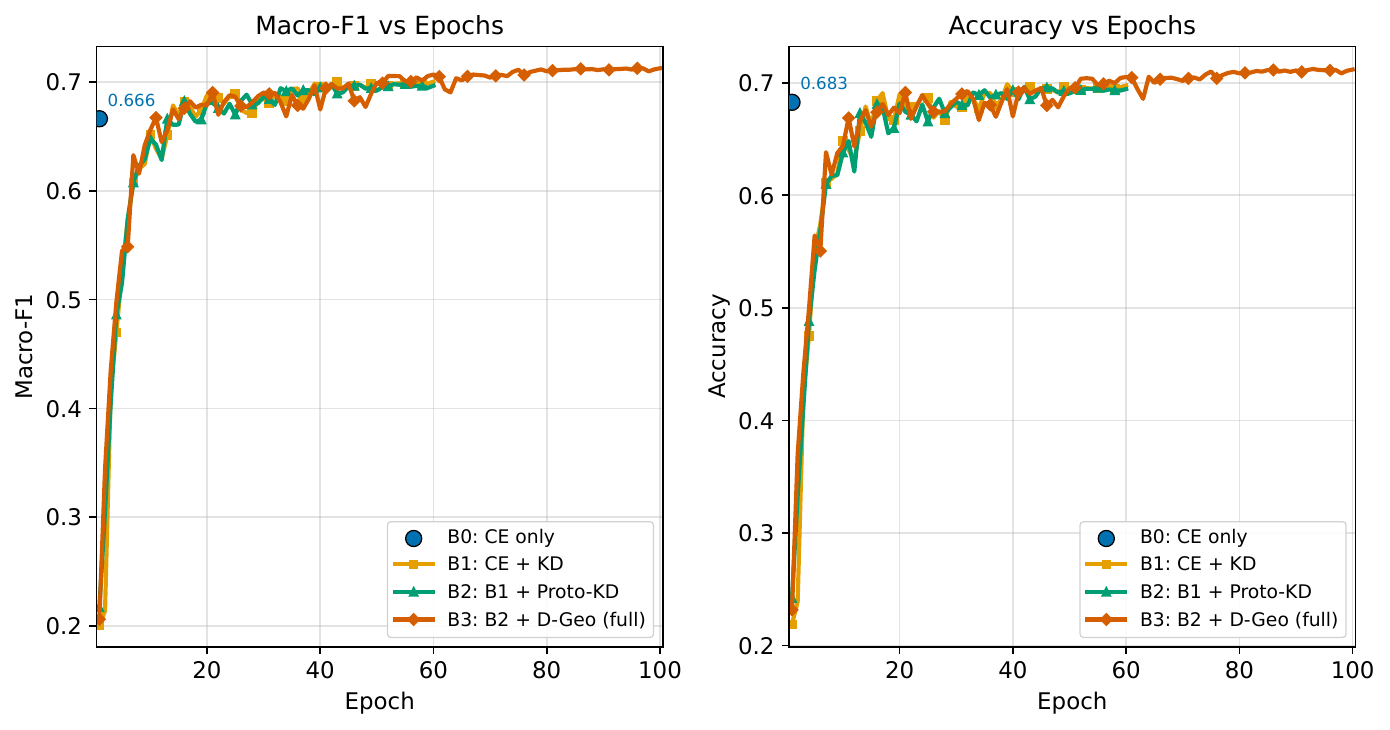}
  \caption{\textbf{Ablation timelines on FERPlus valid.} Macro--F1 (left) and Accuracy (right) vs epochs for
  B0$\!\rightarrow\!$B3. KD (B1) speeds up early convergence, while Proto--KD (B2) and the late-activated D--Geo (B3)
  provide consistent late-epoch gains; B3 attains the best final Macro--F1/Acc.}
  \label{fig:abla_timeline}
\end{figure}

\subsection{Qualitative analysis}
\label{sec:qual}
\textbf{Confusions.} Present-only confusion matrices on AffectNet-mini show reduced anger/sadness swaps with Proto--KD,
and D--Geo further improves high-valence purity.
\textbf{Feature geometry.} t-SNE/UMAP of the 256-D features shows clearer inter-class margins with Proto--KD and
a more compact high-valence cluster with D--Geo \cite{vandermaaten2008tsne,mcinnes2018umap}, while low-valence classes remain separable.
\textbf{Topomap exemplars.} Representative EEG topomaps per V/A bin qualitatively match prototype locations.

\begin{figure}[H]
  \centering
  \includegraphics[width=0.58\linewidth]{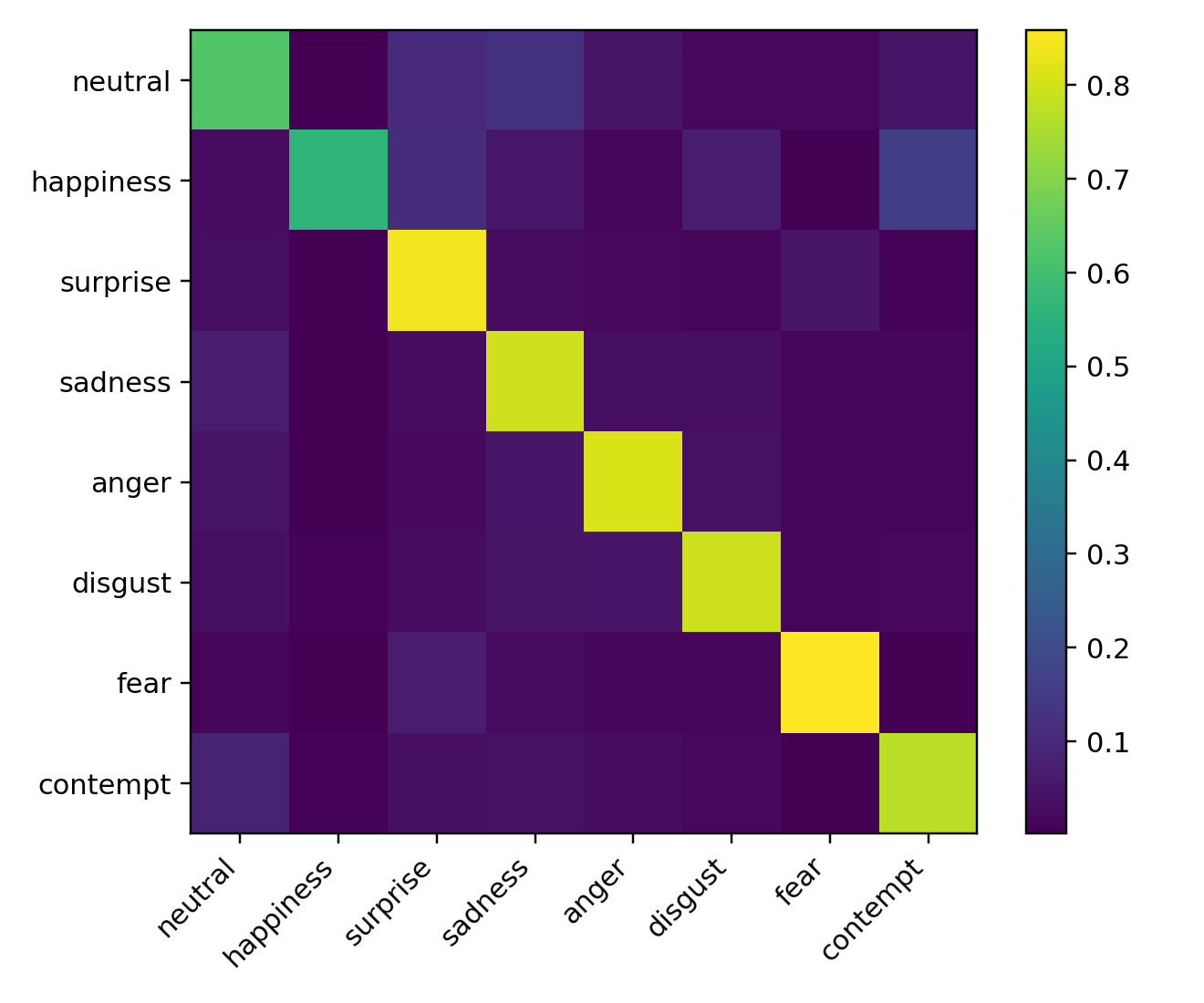}
  \caption{\textbf{Present-only confusion matrix on AffectNet-mini.}
  Evaluated with the full method (\textbf{B3}: CE+KD+Proto--KD+D--Geo).}
  \label{fig:affmini_confmat_present}
\end{figure}

\begin{figure}[H]
  \centering
  \newcommand{\tsneh}{4.2cm}
  \newcommand{\imgshift}{\hspace*{-0.6em}}
  \begin{minipage}[t]{0.24\linewidth}\centering
    \imgshift\includegraphics[height=\tsneh]{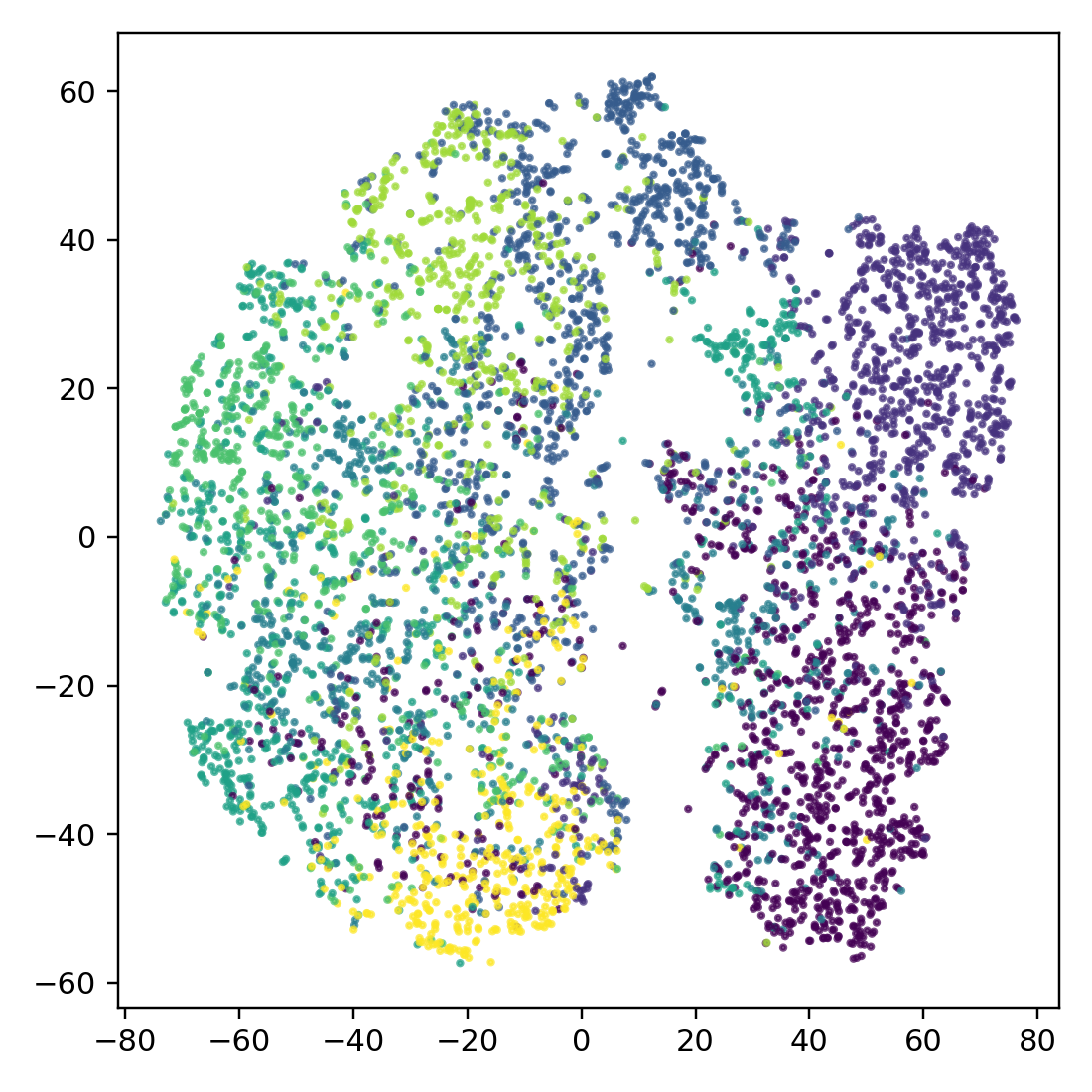}\\[-2pt]
    \footnotesize B0: CE only
  \end{minipage}\hfill
  \begin{minipage}[t]{0.24\linewidth}\centering
    \imgshift\includegraphics[height=\tsneh]{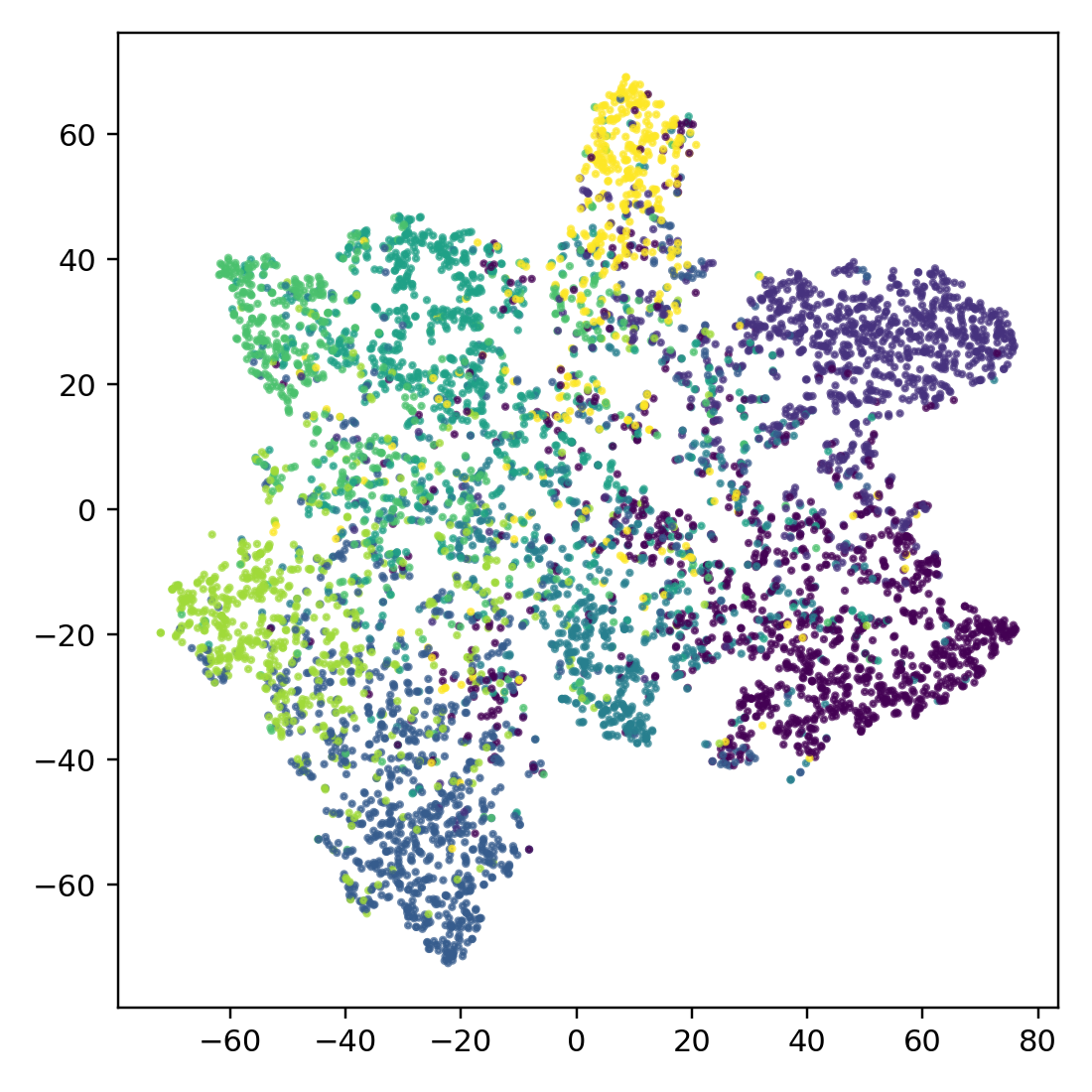}\\[-2pt]
    \footnotesize B1: CE + KD
  \end{minipage}\hfill
  \begin{minipage}[t]{0.24\linewidth}\centering
    \imgshift\includegraphics[height=\tsneh]{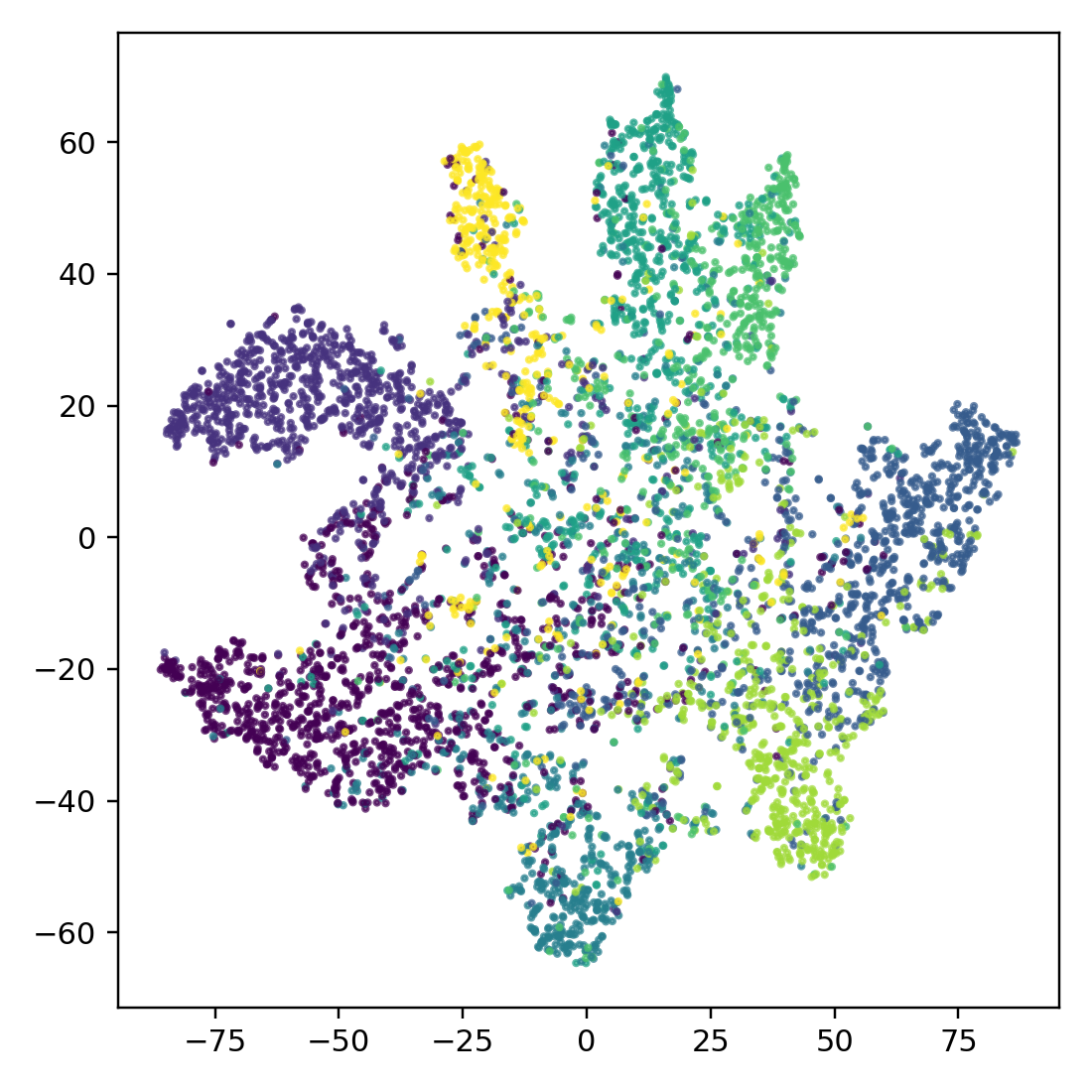}\\[-2pt]
    \footnotesize B2: B1 + Proto--KD
  \end{minipage}\hfill
  \begin{minipage}[t]{0.24\linewidth}\centering
    \imgshift\includegraphics[height=\tsneh]{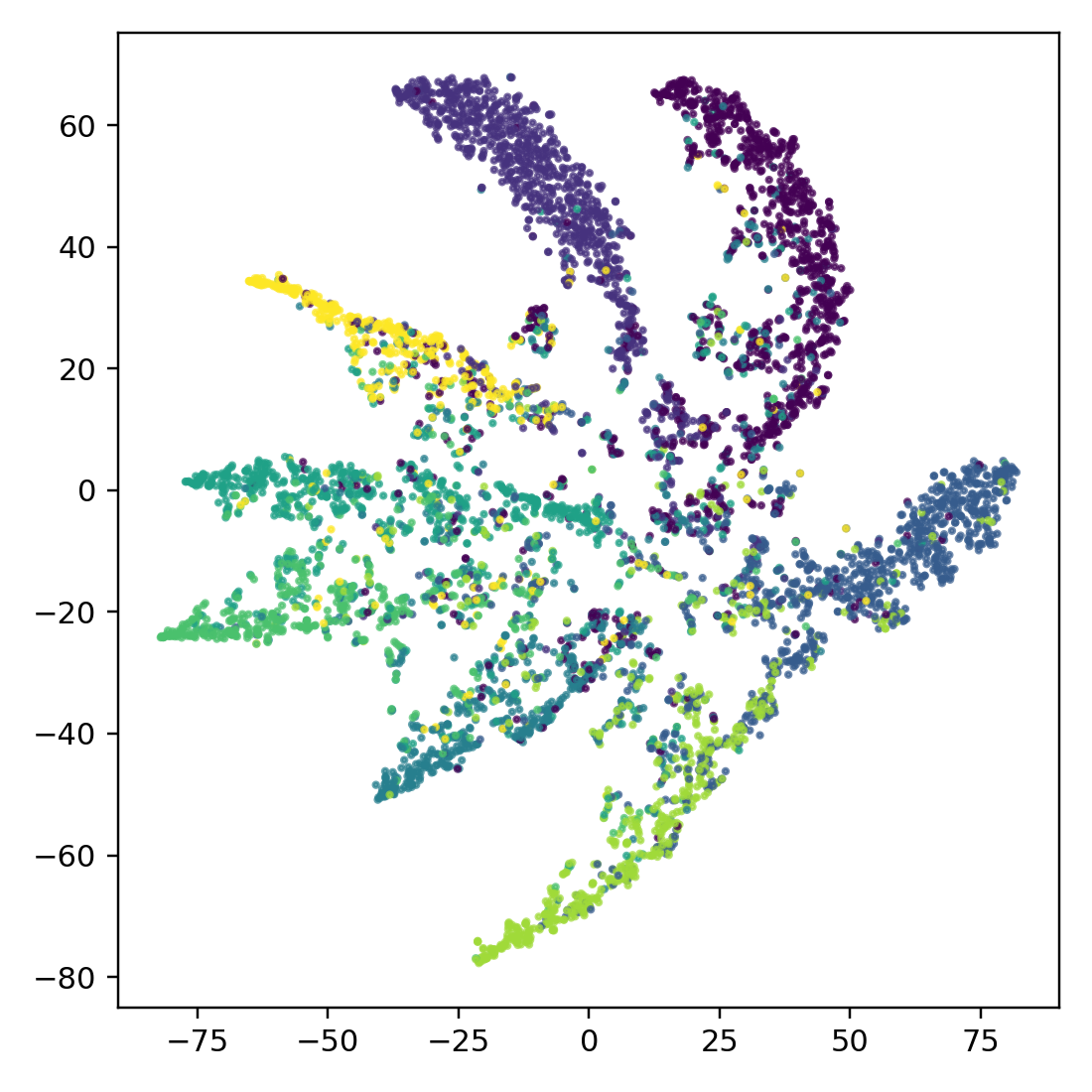}\\[-2pt]
    \footnotesize B3: B2 + D--Geo (full)
  \end{minipage}
  \caption{\textbf{t-SNE of student features on FERPlus valid across ablations (B0$\rightarrow$B3).}
  Proto--KD and D--Geo progressively increase inter-class separability and compactness in high-valence regions \cite{vandermaaten2008tsne,mcinnes2018umap}.}
  \label{fig:tsne_b0_b3}
\end{figure}

\subsection{Training dynamics}
\label{sec:train-dyn}
\noindent\textit{Effect of ablations.} Figure~\ref{fig:abla_timeline} shows that KD (B1) speeds up early convergence \cite{hinton2015distilling},
while Proto--KD (B2) and the late-activated D--Geo (B3) produce consistent late-epoch improvements; B3 attains
the best final Macro--F1/Acc.

\begin{figure}[H]
  \centering
  \hspace{-4mm}
  \includegraphics[width=.98\linewidth]{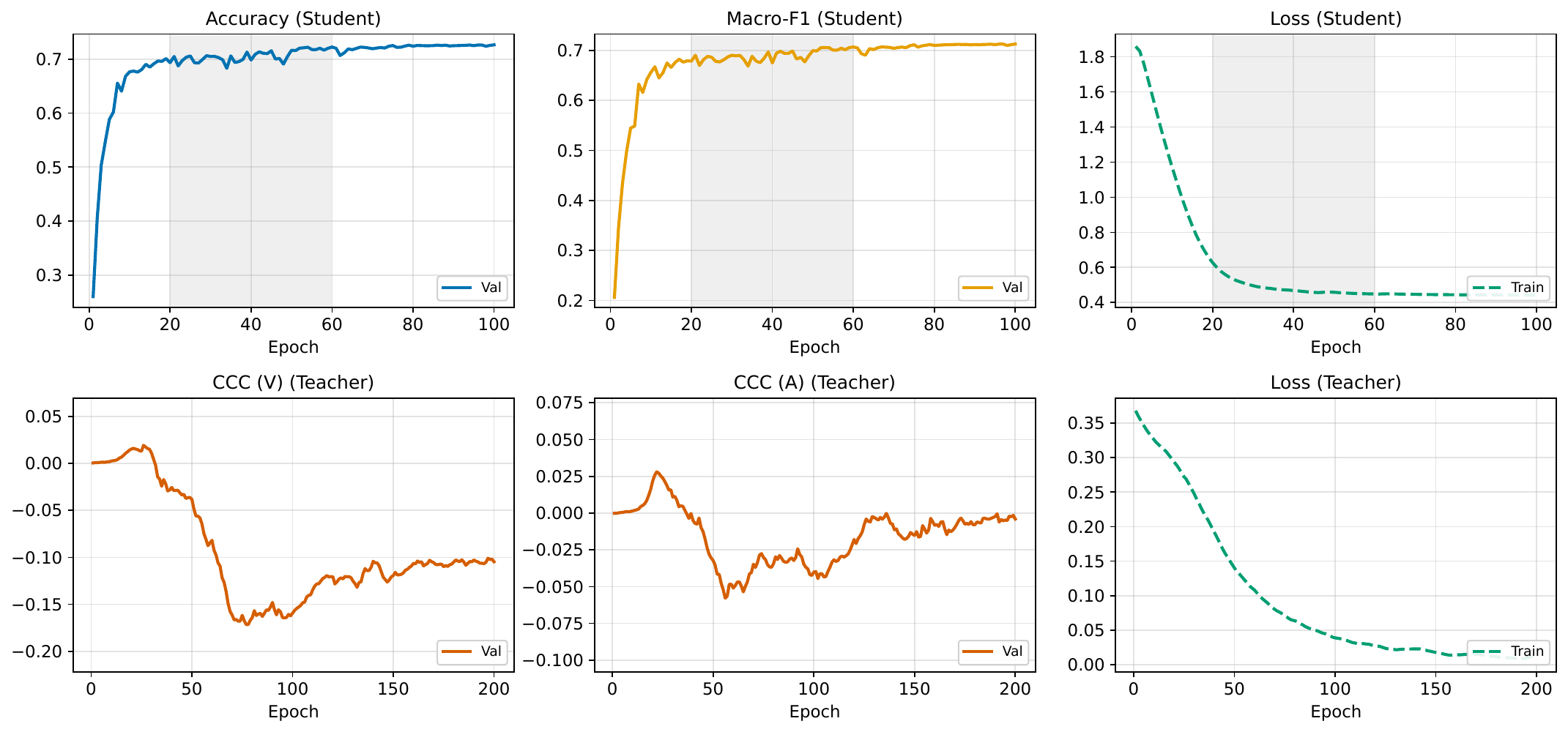}
  \caption{\textbf{Training curves.} Top: student accuracy, Macro--F1 and training loss on FERPlus; 
  Bottom: teacher V/A CCC (val) and training loss. The grey window (epochs 20$\rightarrow$60) marks the 
  late activation period for D--Geo in our recipe (Sec.~\ref{sec:losses}).}
  \label{fig:training_curves}
\end{figure}

\section{Depression-Inspired Prior: Rationale, Scope, and Cautions}
\label{sec:dgeo}

\paragraph{Rationale.}
A large body of affective research associates depressive symptoms with blunted positive affect (anhedonia) \cite{pizzagalli2014anhedonia,treadway2011reconceptualizing}.
We encode only the \emph{shape} of this idea as a weak geometric regularizer on the representation: high–valence regions are encouraged to be slightly more compact, while global inter–class margins are preserved.

\paragraph{How it is applied.}
The D--Geo term is \emph{non-diagnostic} and task-agnostic. It never uses clinical labels and produces no clinical signal.
At inference time the student is a standard FER model; no EEG or clinical attribute is required.
The prototype bank used elsewhere is frozen and anonymized (checksums in Table~\ref{tab:sha256}).

\paragraph{Observed effect (empirical).}
Ablations in Sec.~\ref{sec:ablations} and Fig.~\ref{fig:tsne_b0_b3} show tighter high-valence clusters and
small but consistent late-epoch gains in Macro-F1/Acc when D--Geo is added on top of CE+KD+Proto-KD.
Figure~\ref{fig:abla_timeline} further indicates that KD accelerates early learning, while Proto-KD and the late-activated D--Geo improve the final performance.

\paragraph{Intended scope.}
Our goal is \emph{brain-informed representation shaping for FER}, not mental-health assessment.
We report only FER metrics (Accuracy, Macro-F1, bACC) and release models/prototypes strictly for research on robustness and transfer.

\paragraph{Cautions and limitations.}
\begin{itemize}
  \item \textbf{No clinical use.} The method does not estimate depression and must not be used for screening, triage, or any medical decision.
  \item \textbf{Distribution shift.} Cross-dataset differences (culture, pose, label taxonomies) can interact with regularization; always report per-class and present-only metrics when transferring.
  \item \textbf{Tuning sensitivity.} Enabling D--Geo from epoch~0 or increasing its weight can harm separability; a safe fallback is $\lambda_{\mathrm{geo}}{=}0$.
  \item \textbf{Interpretation.} Tighter clusters are a \emph{modeling bias} we introduce for stability; they should not be interpreted as population evidence.
  \item \textbf{Privacy.} Released prototypes are bin-averaged features (no raw EEG) with reproducible digests (Table~\ref{tab:sha256}).
\end{itemize}

\paragraph{Responsible use (checklist).}
When reusing our code or checkpoints: (i) keep the reported D--Geo weight/schedule; (ii) report both present-only and fixed 8-way metrics; (iii) include with/without D--Geo ablations; (iv) avoid any medical framing; and (v) document datasets and consent/usage terms.

\paragraph{Takeaway.}
D--Geo is a small, transparent bias on representation geometry that improves cross-dataset stability (Sec.~\ref{sec:ablations}) while remaining clearly outside diagnostic claims.

\section{Reproducibility and Artifacts (no code release)}
\label{sec:repro}
We do not release source code at submission time.  To support verification without re-training, we provide
\emph{read-only} artifacts sufficient to reproduce all reported tables and figures derived from evaluation:
(i) the fixed prototype bank (\textbf{v4}, 5$\times$5), (ii) the main student checkpoint
(\textbf{A3\_full}, 100 epochs), and (iii) per-dataset metrics JSONs together with the exact \LaTeX{} tables used in the paper.
Filenames and SHA-256 digests are listed in Table~\ref{tab:sha256}.

\paragraph{What can be checked without any dataset.}
(1) File integrity via SHA-256 (Table~\ref{tab:sha256}); (2) configuration switches and hashes recorded in the
\emph{ablation fingerprint} JSON shipped with the checkpoint (optimizer, loss weights, KD temperature, D--Geo settings); 
(3) tables \textit{verbatim} by compiling the provided \texttt{viz/xval\_main.tex} and \texttt{viz/ablation\_lite.tex}
which read the metrics JSONs we include.

\paragraph{What requires datasets (optional).}
If reviewers have CK+, AffectNet-mini, or FERPlus locally, the student checkpoint can be evaluated using any standard
inference pipeline that follows our protocol in Sec.~\ref{sec:impl} and Sec.~\ref{sec:exp}
(input size $224$, 8-way fixed mapping for cross-dataset, present-only metrics as defined in Sec.~\ref{sec:protocols}).
This step is \emph{not} required to verify the paper numbers because we already ship the metrics JSONs.

\paragraph{Scope and access.}
We release only model weights, metrics, and tables; we do not redistribute datasets or training code.  
Artifacts are provided for \emph{review-only} use to enable integrity checks and table regeneration. 
Upon publication, we will maintain a stable artifact bundle (weights + metrics + tables) at the camera-ready link.

\section{Limitations and Ethics}
\label{sec:limits-ethics}

\paragraph{Limitations.}
\begin{itemize}[leftmargin=1.1em,itemsep=2pt,topsep=2pt]
  \item \textbf{Non-diagnostic scope.} The depression-inspired geometric prior (D--Geo) is a weak regularizer on representation geometry, not a clinical model.
  \item \textbf{Dataset bias.} Public FER datasets can contain demographic and capture biases. We report cross-dataset results and \emph{present-only} metrics to reduce label-set confounds, but a full fairness audit (e.g., subgroup analysis) is left to future work.
  \item \textbf{Prototype granularity.} Using a fixed $5{\times}5$ V/A grid improves stability but may underfit finer affect dynamics; adaptive or data-driven grids are a promising extension.
  \item \textbf{Teacher dependence.} Prototype quality depends on the teacher trained on EEG topomaps; suboptimal teachers or shifts in V/A calibration may cap the attainable gains.
  \item \textbf{Privacy and deployment.} Physiological signals (EEG/gaze) are used only during development to form static prototypes; the deployed student is vision-only. We do not display any identifiable face exemplars.
\end{itemize}

\paragraph{Ethics \& broader impact.}
We use only publicly available datasets (\textsc{CK+}, \textsc{AffectNet}, \textsc{FERPlus}/\textsc{FER2013}, \textsc{DREAMER}, \textsc{MAHNOB--HCI}) under their academic licenses. Released artifacts for review are limited to model weights, metrics, and tables; they contain no personally identifiable images.

\section{Conclusion}
We presented \textsc{NeuroGaze–Distill}, a brain-informed yet deployment-friendly framework for facial expression recognition (FER). The method couples a \emph{static} EEG-derived prototype bank with a lightweight \emph{depression-inspired} geometric prior (D–Geo), and distills both cues into a conventional CNN student via logit and prototype matching. The prototype bank (v4; DREAMER with MAHNOB–HCI as a stability supplement) is formed once on teacher features, frozen thereafter, and—crucially—requires \emph{no} paired EEG–face data and \emph{no} non-visual signals at inference. Students remain standard backbones (ResNet-18/50) \cite{he2016deep} trained with CE+KD+Proto-KD+D–Geo using modest hyperparameters (e.g., $T{=}5$, $\tau{=}0.90$).

\paragraph{Empirical takeaways.}
Across datasets (Sec.~\ref{sec:exp}), the full model (\textbf{B3}) improves Accuracy, Macro-F1, and bACC over the CE baseline and intermediate variants (Tables~\ref{tab:xval_main} and \ref{tab:abla_lite}). The ablation timelines (Fig.~\ref{fig:abla_timeline}) show a clear division of labor: KD (\textbf{B1}) accelerates early learning \cite{hinton2015distilling}, while Proto-KD (\textbf{B2}) and the late-activated D--Geo (\textbf{B3}) yield consistent late-epoch gains. Qualitatively, t-SNE/UMAP panels (Fig.~\ref{fig:tsne_b0_b3}) reveal tighter, better-separated clusters—especially for high-valence categories—without sacrificing low-valence separability \cite{vandermaaten2008tsne,mcinnes2018umap}.

\paragraph{Practicality and verification.}
The recipe uses off-the-shelf training components (AdamW \cite{loshchilov2019adamw}, cosine schedule \cite{loshchilov2017sgdr}, AMP, channels-last, light clipping) and a single frozen prototype bank shared by all students and datasets. We provide SHA-256–versioned artifacts—prototype bank, the main student checkpoint, and metrics bundles—summarized in Table~\ref{tab:sha256}.

\paragraph{Outlook.}
Promising directions include: (i) adaptive/data-driven prototype granularity beyond a fixed $5{\times}5$ V/A grid; (ii) teacher calibration and stronger students (e.g., ViT backbones \cite{dosovitskiy2021vit}) while keeping inference vision-only; (iii) broader fairness analyses across demographics and capture conditions; and (iv) additional \emph{privileged} teachers (speech or physiology) during training under appropriate governance.

% =========================
% References (paste before \end{document})
% =========================

\appendix
\section{Appendix}

\subsection{Metrics and confidence intervals}
\label{app:metrics}
We report Accuracy (Acc), Macro-F1, balanced Accuracy (bACC) \cite{brodersen2010balanced}, and optionally Macro-AUROC.
Let $\mathcal{C}$ be the class set, $n_c$ the number of samples in class $c$, $\mathrm{TP}_c$ true positives,
$\mathrm{TPR}_c$ true positive rate, and $\mathrm{F1}_c$ the per-class F1.
\[
\mathrm{Acc}=\frac{1}{\sum_{c} n_c}\sum_{c}\mathrm{TP}_c,\qquad
\mathrm{Macro\text{-}F1}= \frac{1}{|\mathcal{C}|}\sum_{c\in\mathcal{C}}\mathrm{F1}_c,\qquad
\mathrm{bACC}= \frac{1}{|\mathcal{C}|}\sum_{c\in\mathcal{C}}\mathrm{TPR}_c.
\]
For Macro-AUROC we compute one-vs-rest AUROC for each class and average across classes.
Confidence intervals are obtained by stratified bootstrapping over samples with 1{,}000 resamples
and a fixed random seed; the 2.5 and 97.5 percentiles form the 95\% interval \cite{efron1994bootstrap}.

\subsection{Cross-dataset protocols}
\label{app:protocols}
\textbf{Within-domain.} FERPlus validation split.  
\textbf{Cross-dataset.} A student trained on FERPlus is evaluated on AffectNet-mini
(and optionally CK+ \cite{lucey2010ckplus}). We report two settings:
(i) \emph{8-way fixed mapping} using a consistent FER mapping across datasets; and
(ii) \emph{present-only}, where metrics are computed only over the set of labels that exist
in the target dataset annotations (Sec.~\ref{sec:protocols}).

\subsection{Training recipe (consolidated)}
\label{app:recipe}
Unless stated otherwise, all ablations share the same backbone and recipe.
\begin{itemize}[leftmargin=*, itemsep=2pt, topsep=2pt]
  \item \textbf{Backbone:} ResNet-18/50 with a 256-D projection and an 8-way classifier \cite{he2016deep}.
  \item \textbf{Optimization:} AdamW \cite{loshchilov2019adamw}, cosine schedule \cite{loshchilov2017sgdr}, base learning rate $2\times10^{-4}$,
        weight decay $0.05$, batch size 128, mixed precision (AMP), channels-last memory format,
        gradient clipping at 1.0, label smoothing $\alpha=0.055$ \cite{muller2019labelsmoothing}, and mild class weights.
  \item \textbf{Logit KD:} temperature $T=5.0$ with MSE/KL objective; medium loss weight \cite{hinton2015distilling}.
  \item \textbf{Prototype KD:} cosine similarity, feature temperature $\tau=0.90$; small loss weight（cf. \cite{snell2017prototypical,movshovitz2017proxynca}）.
  \item \textbf{D-Geo:} small weight with a late cosine schedule（epochs $20\rightarrow60$）; positives set $\{\text{happiness}, \text{surprise}\}$ \cite{pizzagalli2014anhedonia,treadway2011reconceptualizing}.
  \item \textbf{Teacher prototypes:} fixed v4 bank formed on DREAMER with MAHNOB-HCI for stability \cite{katsigiannis2018dreamer,soleymani2012mahnob}.
\end{itemize}

\subsection{Artifacts for verification}
\label{app:artifacts}
We release only non-identifiable artifacts (no images) sufficient to verify the reported numbers.
\begin{itemize}[leftmargin=*, itemsep=2pt, topsep=2pt]
  \item \textbf{Main checkpoint:} \texttt{outs/abla\_A3\_full\_100/student\_best.ckpt}.
  \item \textbf{Per-dataset metrics:} \texttt{outs/xval\_ferplus\_valid\_abla\_A3\_full/metrics.json},\\
        \texttt{outs/xval\_ckplus\_abla\_A3\_full/metrics.json},\\
        \texttt{outs/xval\_affmini\_abla\_A3\_full/metrics.json}.
  \item \textbf{Ready-to-compile tables:} \texttt{viz/xval\_main.tex} and \texttt{viz/ablation\_lite.tex}.
  \item \textbf{Configuration fingerprint:} \texttt{outs/abla\_A3\_full\_100/ablation\_fingerprint.json}.
  \item \textbf{Checksums:} \texttt{SHA256SUMS.txt}. Verify with \texttt{sha256sum -c SHA256SUMS.txt}.
  \item \textbf{Released files and digests:} see Table~\ref{tab:sha256}.
\end{itemize}

\subsection{Notes on limitations of the package}
\label{app:limits}
The package does not redistribute any dataset media and does not include training code.
It is intended for auditing the numbers reported in the paper and for re-emitting
the same LaTeX tables from the shipped JSON metrics when the datasets are unavailable.


\begin{thebibliography}{99}

\bibitem{he2016deep}
Kaiming He, Xiangyu Zhang, Shaoqing Ren, and Jian Sun.
Deep residual learning for image recognition.
In \emph{CVPR}, pp. 770--778, 2016.

\bibitem{dosovitskiy2021vit}
Alexey Dosovitskiy, Lucas Beyer, Alexander Kolesnikov, et al.
An Image is Worth 16x16 Words: Transformers for Image Recognition at Scale.
In \emph{ICLR}, 2021. arXiv:2010.11929.

\bibitem{kingma2015adam}
Diederik P. Kingma and Jimmy Ba.
Adam: A Method for Stochastic Optimization.
In \emph{ICLR}, 2015. arXiv:1412.6980.

\bibitem{loshchilov2019adamw}
Ilya Loshchilov and Frank Hutter.
Decoupled Weight Decay Regularization.
In \emph{ICLR}, 2019. arXiv:1711.05101.

\bibitem{loshchilov2017sgdr}
Ilya Loshchilov and Frank Hutter.
SGDR: Stochastic Gradient Descent with Warm Restarts.
In \emph{ICLR}, 2017. arXiv:1608.03983.

\bibitem{hinton2015distilling}
Geoffrey Hinton, Oriol Vinyals, and Jeff Dean.
Distilling the Knowledge in a Neural Network.
\emph{NeurIPS Deep Learning Workshop}, 2015. arXiv:1503.02531.

\bibitem{bucilua2006model}
Cristian Bucilu{\u{a}}, Rich Caruana, and Alexandru Niculescu-Mizil.
Model Compression.
In \emph{KDD}, pp. 535--541, 2006.

\bibitem{muller2019labelsmoothing}
Rafael M{\"u}ller, Simon Kornblith, and Geoffrey Hinton.
When Does Label Smoothing Help?
In \emph{NeurIPS}, 2019.

\bibitem{szegedy2016rethinking}
Christian Szegedy, Vincent Vanhoucke, Sergey Ioffe, Jon Shlens, and Zbigniew Wojna.
Rethinking the Inception Architecture for Computer Vision.
In \emph{CVPR}, pp. 2818--2826, 2016.

\bibitem{tarvainen2017meanteacher}
Antti Tarvainen and Harri Valpola.
Mean Teachers are Better Role Models: Weight-Averaged Consistency Targets Improve Semi-Supervised Deep Learning.
In \emph{NeurIPS}, 2017. arXiv:1703.01780.

\bibitem{vandermaaten2008tsne}
Laurens van der Maaten and Geoffrey Hinton.
Visualizing Data using t-SNE.
\emph{Journal of Machine Learning Research}, 9:2579--2605, 2008.

\bibitem{mcinnes2018umap}
Leland McInnes, John Healy, and James Melville.
UMAP: Uniform Manifold Approximation and Projection for Dimension Reduction.
arXiv:1802.03426, 2018.

\bibitem{snell2017prototypical}
Jake Snell, Kevin Swersky, and Richard S. Zemel.
Prototypical Networks for Few-shot Learning.
In \emph{NeurIPS}, 2017.

\bibitem{movshovitz2017proxynca}
Yair Movshovitz-Attias, Alexander Toshev, Thomas Leung, Sergey Ioffe, and Saurabh Singh.
No Fuss Distance Metric Learning using Proxies.
In \emph{ICCV}, pp. 360--368, 2017.

\bibitem{lucey2010ckplus}
Patrick Lucey, Jeffrey F. Cohn, Takeo Kanade, Jason Saragih, Zara Ambadar, and Iain Matthews.
The Extended Cohn-Kanade Dataset (CK+): A complete dataset for action unit and emotion-specified expression.
In \emph{CVPR Workshops}, pp. 94--101, 2010.

\bibitem{mollahosseini2019affectnet}
Ali Mollahosseini, Behzad Hasani, and Mohammad H. Mahoor.
AffectNet: A Database for Facial Expression, Valence, and Arousal Computing in the Wild.
\emph{IEEE Transactions on Affective Computing}, 10(1):18--31, 2019.
doi:10.1109/TAFFC.2017.2740923.

\bibitem{barsoum2016ferplus}
Emad Barsoum, Cha Zhang, Cristian Canton Ferrer, and Zhifeng Zhang.
Training Deep Networks for Facial Expression Recognition with Crowd-Sourced Label Distribution.
In \emph{ACM ICMI}, pp. 279--283, 2016.
doi:10.1145/2993148.2993165.

\bibitem{soleymani2012mahnob}
Mohammad Soleymani, Jeroen Lichtenauer, Florian Eyben, Markus K{\"a}chele, et al.
MAHNOB-HCI: A Multimodal Database for Affect Recognition and Implicit Tagging.
\emph{IEEE Transactions on Affective Computing}, 3(1):90--102, 2012.
doi:10.1109/T-AFFC.2011.34.

\bibitem{katsigiannis2018dreamer}
Stamatios Katsigiannis and Naeem Ramzan.
DREAMER: A Database for Emotion Recognition through EEG and ECG Signals from Wireless Low-cost Off-the-Shelf Devices.
\emph{IEEE Journal of Biomedical and Health Informatics}, 22(1):98--107, 2018.
doi:10.1109/JBHI.2017.2688239.

\bibitem{russell1980circumplex}
James A. Russell.
A Circumplex Model of Affect.
\emph{Journal of Personality and Social Psychology}, 39(6):1161--1178, 1980.

\bibitem{pizzagalli2014anhedonia}
Diego A. Pizzagalli.
Depression, Stress, and Anhedonia: Toward a Synthesis and Integrated Model.
\emph{Annual Review of Clinical Psychology}, 10:393--423, 2014.
doi:10.1146/annurev-clinpsy-050212-185606.

\bibitem{treadway2011reconceptualizing}
Michael T. Treadway and David H. Zald.
Reconceptualizing Anhedonia: Novel Perspectives on Balancing the Anticipation and Experience of Reward.
\emph{Psychological Bulletin}, 137(6):1085--1108, 2011.
doi:10.1037/a0024515.

\bibitem{efron1994bootstrap}
Bradley Efron and Robert J. Tibshirani.
\emph{An Introduction to the Bootstrap}.
Chapman and Hall/CRC, 1994.

\bibitem{brodersen2010balanced}
Kay H. Brodersen, Cheng Soon Ong, Klaas E. Stephan, and Joachim M. Buhmann.
The balanced accuracy and its posterior distribution.
In \emph{ICPR}, pp. 3121--3124, 2010.

\end{thebibliography}
\end{document}